\documentclass{article}

\usepackage{PRIMEarxiv}

\usepackage[utf8]{inputenc} 
\usepackage[T1]{fontenc}    
\usepackage{hyperref}       
\usepackage{url}            
\usepackage{booktabs}       
\usepackage{amsfonts}       
\usepackage{nicefrac}       
\usepackage{microtype}      
\usepackage{lipsum}
\usepackage{amssymb,amsthm,amsmath}
\usepackage{graphicx}
\graphicspath{{media/}}     
\usepackage{booktabs}%
\usepackage{algorithm}%
\usepackage{algorithmicx}%
\usepackage{algpseudocode}%
\usepackage{listings}%
\usepackage{tikz}
\usepackage{subcaption}
\usetikzlibrary{positioning, shapes.geometric, arrows.meta}

\title{Quantum-Inspired DRL Approach with LSTM and OU Noise for Cut Order Planning Optimization
\thanks{\textit{\underline{Citation}}: 
\textbf{Chrisnanto Y.H., Chrisnanto J.E. Quantum-Inspired DRL Approach with LSTM and OU Noise for Cut Order Planning Optimization. Pages.... DOI:000000/11111.}} 
}

\author{
  Yulison Herry Chrisnanto \\
  Department of Informatics \\
  Jenderal of Achmad Yani University \\
  Cimahi\\
  \texttt{yhc@if.unjani.ac.id} \\
   \And
  Julian Evan Chrisnanto \\
  Department of Physics \\
  Universitas Padjadjaran \\
  Jatinangor\\
  \texttt{julian20001@mail.unpad.ac.id} \\
}

\begin{document}
\maketitle

\begin{abstract}
Cut order planning (COP) is a critical challenge in the textile industry, directly impacting fabric utilization and production costs. Conventional methods based on static heuristics and catalog-based estimations often struggle to adapt to dynamic production environments, resulting in suboptimal solutions and increased waste. In response, we propose a novel Quantum-Inspired Deep Reinforcement Learning (QI-DRL) framework that integrates Long Short-Term Memory (LSTM) networks with Ornstein-Uhlenbeck noise. This hybrid approach is designed to explicitly address key research questions regarding the benefits of quantum-inspired probabilistic representations, the role of LSTM-based memory in capturing sequential dependencies, and the effectiveness of OU noise in facilitating smooth exploration and faster convergence. Extensive training over 1000 episodes demonstrates robust performance, with an average reward of 0.81 (±0.03) and a steady decrease in prediction loss to 0.15 (±0.02). A comparative analysis reveals that the proposed approach achieves fabric cost savings of up to 13\% compared to conventional methods. Furthermore, statistical evaluations indicate low variability and stable convergence. Despite the fact that the simulation model makes several simplifying assumptions, these promising results underscore the potential of the scalable and adaptive framework to enhance manufacturing efficiency and pave the way for future innovations in COP optimization.
\end{abstract}


\section{Introduction}\label{sec1}

Cut Order Planning (COP) is crucial in the textile and manufacturing industries for cost management, efficiency, and adaptability to market demands. By optimizing COP, manufacturers can significantly reduce material waste—accounting for 50-60\% of total manufacturing costs—while enhancing fabric utilization and production performance. However, challenges arise from increasing customization and fragmented orders, complicating efficient planning. Furthermore, fabric constraints and the need to balance cost reduction with production timelines add further complexity \cite{wong_optimizing_2013}. Implementing advanced optimization models for Cut Order Planning (COP) requires significant technological resources, which may not be accessible to all manufacturers. The dynamic nature of the fashion industry adds complexity, demanding adaptability while maintaining stable and efficient planning. COP must balance fabric efficiency with labor and machine costs, as intricate garment patterns can increase operational expenses. In addition, rapid fashion trends require quick response strategies, further complicating computational processes. Although automation can improve efficiency, its implementation is time-consuming, making it challenging for manufacturers to optimize production while minimizing waste \cite{wong_genetic_2008}. Production planning and scheduling must address key constraints, including fixed board length, marker length, and demand distribution. The board length constraint ensures tasks do not exceed a set maximum, optimizing resource use, while the marker length constraint defines the space required for task representation. Demand distribution ensures production aligns with required quantities over time. To manage these complexities, adaptive optimization techniques are essential, integrating multi-level algorithms, machine learning for demand forecasting, hybrid optimization methods, and constraint programming. This approach enhances resource allocation, accommodates demand fluctuations, and improves overall operational efficiency \cite{kis_cutting_2012}.

The study in \cite{tsao_marker_2022} addresses the marker planning problem in apparel manufacturing by using a pixel-based representation for irregular patterns and a Moving Heuristic (MH) combined with a hybrid Particle Swarm Optimization (PSO) framework enhanced by local search, Genetic Algorithm (GA), and Simulated Annealing (SA) to optimize pattern arrangements and reduce fabric usage. However, its computational intensity and limitation to 0° and 180° rotation angles may lead to extended processing times and reduced adaptability for complex fabric patterns. The study in \cite{mhallah_heuristics_2016}, introduces an introduces an innovative approach by treating cut order planning (COP) and two-dimensional layout (TDL) as a unified optimization problem (CT = COP + TDL), rather than addressing them separately. The authors employ constructive heuristics and various meta-heuristic techniques, including stochastic local improvement methods, genetic algorithms, and a hybrid approach, to minimize total fabric length. This approach enhances fabric utilization and reduces waste.Their computational experiments on benchmark instances and industrial cases demonstrate that this integrated strategy can significantly improve efficiency and accuracy in fabric usage. This is particularly relevant given that fabric costs constitute a major portion of overall expenses in apparel manufacturing. However, the study acknowledges the NP-complete nature of the problem, which poses challenges for exact solutions. Furthermore, the scalability and practical implementation of these heuristics may be limited in more complex, real-world scenarios. The study in \cite{arbib_onedimensional_2016} address the one-dimensional cutting stock problem by proposing a novel integer linear programming formulation that imposes limits on the number of different part types produced concurrently, thereby regulating open stacks.Their model, which exhibits a quadratic growth in constraints relative to the number of part types, employs a standard column generation approach to efficiently solve problems with up to 20 distinct part types—instances where optimal solutions were previously uncharted. While the formulation proves computationally viable for smaller instances, it may encounter challenges with larger problems due to the exponential growth of variables, with heuristic solutions sometimes slightly exceeding optimal values and memory limitations emerging during solver use (e.g., with CPLEX). The study in \cite{gahm_applying_2022} enhances hierarchical production planning by integrating machine learning to improve complex nesting solutions and employing a rigorous instance generation procedure based on real-world technical drawings. However, its reliance on ML approximations may overlook some nesting configurations, and focusing on specific dimensions like height may oversimplify the complexity of nesting relationships, potentially reducing feasibility accuracy. The study in \cite{zhang_heuristic_2024}, introduces the Two-Step Heuristic (TSH) and Node Routing Method (NRM) for the cutting path problem, optimizing two-dimensional layouts by enhancing efficiency and precision—yielding optimal solutions for small instances and near-optimal solutions for larger ones. However, the heuristic nature of the methods means they do not guarantee optimal solutions for complex or large instances, and their performance may be limited by unaddressed practical constraints. The study in \cite{deza_machine_2023} provides a comprehensive analysis of recent advancements in applying machine learning techniques—such as reinforcement, imitation, and supervised learning—to enhance cut selection in Mixed-Integer Linear Programming (MILP), addressing tasks like cut generation, scoring, and decision-making during the branch and bound process. However, the study highlights significant limitations including the lack of fair baselines, large-scale datasets, and a unified framework, which challenge the practical implementation and generalization of these methods in real-world MILP solvers. The study \cite{wang_learning_2024} introduces Hierarchical Sequence/Set Model (HEM), a novel hierarchical model that uses reinforcement learning to optimize cut selection in MILP problems, achieving significant efficiency improvements, including up to 91.29\% and 97.72\% gains in time and primal-dual gap metrics. However, its performance may plateau due to challenges in exploring a vast action space, complex interactions between hierarchical levels, and difficulties in accurately determining the optimal number of cuts for diverse problem instances. In study \cite{ranaweera_optimal_2023}, provides a comprehensive review of heuristic and meta-heuristic approaches for cut order planning, highlighting the dominance of genetic algorithms and their hybridization in minimizing fabric waste and improving efficiency. However, the study primarily focuses on fabric utilization while overlooking labor costs and energy consumption, lacks an in-depth evaluation of computational scalability, and relies on simulated data rather than real-world validation. In study \cite{wang_learning_2023}, introduces a Hierarchical Sequence Model (HEM) using a sequence-to-sequence learning paradigm with attention mechanisms to optimize cut selection and ordering, significantly improving Mixed-Integer Linear Programming (MILP) solver performance. However, its reliance on a pointer network may limit generalization in complex or large-scale problems, and further evaluation is needed to assess its scalability and robustness across diverse MILP instances. The proposed method in \cite{choi_reinforcement_2023}, integrates reinforcement learning with an attention mechanism to optimize cut and fill operations dynamically, enhancing adaptability, route efficiency, and cost reduction while generalizing across scenarios without extensive re-optimization. However, the model may produce suboptimal policies in repetitive layouts, faces uncertainty in entirely new scenarios, and requires significant computational resources, limiting its feasibility for large-scale applications. The study \cite{cals_solving_2021}, applies Deep Reinforcement Learning (DRL) to the Order Batching and Sequencing Problem (OBSP) in hybrid warehouse systems, using Proximal Policy Optimization (PPO) to adaptively optimize order batching and sequencing, outperforming traditional heuristics in dynamic environments. However, the model faces limitations in computational demands, generalization to new warehouse configurations, interpretability of decisions, and sensitivity to hyperparameter tuning, requiring extensive experimentation for optimal performance.

Despite significant advancements in cut order planning through heuristic, machine learning, and reinforcement learning approaches, extant methods continue to confront limitations in computational efficiency, adaptability, and scalability. Traditional heuristics, including Genetic Algorithms and Particle Swarm Optimization, have exhibited effectiveness; however, they frequently encounter challenges in terms of real-time adaptability and large-scale industrial applications. Likewise, reinforcement learning techniques applied to related optimization problems, including order batching and cut selection, have shown promise but are hindered by computational demands, limited generalization, and reliance on extensive training data. Given these challenges, there is a need for a more adaptive and efficient optimization framework that can dynamically learn optimal cut order planning strategies while balancing waste minimization and production efficiency.

To address these limitations, we propose a Quantum-Inspired Deep Reinforcement Learning (QI-DRL) model \cite{li_deep_2018, li_quantum-inspired_2024} that integrates Long Short-Term Memory (LSTM) networks \cite{lindemann_survey_2021} for sequential decision-making with Ornstein-Uhlenbeck (OU) noise for enhanced exploration \cite{carter_parameter_2024}. Our approach harnesses quantum-inspired principles—such as the probabilistic interpretation inherent in quantum superposition—to improve action selection, enabling the model to explore a broader solution space efficiently and avoid premature convergence. The incorporation of LSTM ensures that the model retains memory of previous optimization states, thereby facilitating more informed decisions across iterations, while OU noise introduces controlled, temporally correlated perturbations that promote smoother exploration. Through this hybrid framework, our goal is to enhance fabric utilization, minimize waste, and optimize production balance in a dynamic and scalable manner.

In summary, the introduction establishes the critical importance of cut order planning (COP) in the apparel industry, emphasizing its role in reducing fabric waste and production costs.Traditional methods often rely on independent processing of COP and two-dimensional layout (TDL), leading to estimation errors and inefficiencies. The proposed approach, which integrates quantum-inspired deep reinforcement learning (QI-DRL) with LSTM networks and Ornstein-Uhlenbeck noise, overcomes these limitations by capturing sequential dependencies and enabling smoother exploration of the solution space.This innovative framework facilitates a comprehensive exploration of improved fabric utilization and enhanced production efficiency in garment manufacturing.

\section{Materials and Method}\label{sec2}

This section delineates the framework and experimental setup employed to address the COP problem.The cut order planning challenge is formulated as a sequential decision-making task with clearly defined objective functions and operational constraints.The approach integrates quantum-inspired deep reinforcement learning with LSTM networks to capture long-term dependencies and employs Ornstein-Uhlenbeck noise for smooth exploration. The section also describes the simulation environment, parameter tuning, and evaluation metrics—such as average reward and loss—to ensure robust and reproducible results, setting a solid foundation for our subsequent analysis.

In order to provide a clear framework for the research's objectives, the following research questions have been articulated:
\begin{enumerate}
    \item Does the integration of quantum-inspired probabilistic representations with deep reinforcement learning enhance decision-making in cut order planning?
    \item Does the incorporation of Long Short-Term Memory (LSTM) networks enhance the model's capacity to capture the sequential dependencies that are inherent in production process?
    \item How effective is the use of Ornstein-Uhlenbeck noise in promoting smooth exploration and accelerating the convergence of the reinforcement learning framework?
    \item Can the proposed Quantum-Inspired Deep Reinforcement Learning (QI-DRL) model achieve significant reductions in fabric waste while reliably meeting production demands compared to traditional optimization methods?
\end{enumerate}

These inquiries form the foundation of our investigation into the development of a robust, scalable solution that addresses both the operational constraints and the dynamic requirements of modern garment manufacturing.

\subsection{Cut Order Planning}
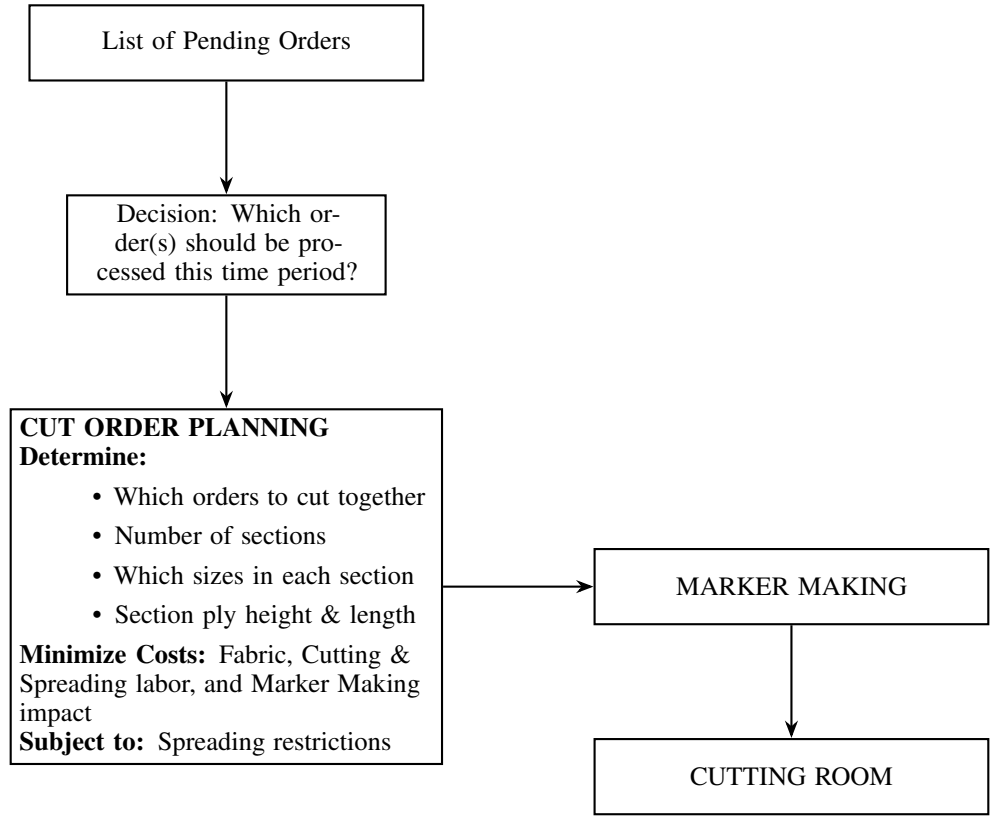
\begin{figure}[htbp]
\noindent\hspace*{3cm}
\begin{tikzpicture}[
    node distance=1.5cm,
    box/.style={rectangle, draw, thick, text width=5cm, minimum height=1cm, align=center},
    bigbox/.style={rectangle, draw, thick, text width=5.5cm, minimum height=3.5cm, align=left},
    arrow/.style={-Stealth, thick}
]

\node[box] (pending) {List of Pending Orders};

\node[box, below=of pending, text width=4cm] (decision) {Decision: Which order(s) should be processed this time period?};

\node[bigbox, below=of decision] (planning) {
    \textbf{CUT ORDER PLANNING}
    
    \textbf{Determine:}
    \begin{itemize}
        \item Which orders to cut together
        \item Number of sections
        \item Which sizes in each section
        \item Section ply height \& length
    \end{itemize}
    \textbf{Minimize Costs:} Fabric, Cutting \& Spreading labor, and Marker Making impact
    
    \textbf{Subject to:} Spreading restrictions
};

\node[box, right=2cm of planning] (marker) {MARKER MAKING};

\node[box, below=of marker] (cutting) {CUTTING ROOM};

\draw[arrow] (pending) -- (decision);
\draw[arrow] (decision) -- (planning);
\draw[arrow] (planning) -- (marker);
\draw[arrow] (marker) -- (cutting);
\end{tikzpicture}
\caption{General process of COP. \cite{jacobs-blecha_cut_1998}}\label{fig1}
\end{figure}

Cut Order Planning (COP) is a foundational process in the domain of apparel manufacturing. It ensures the efficient production of fabric by determining the optimal method for its cutting to align with customer orders, thereby minimizing overall costs. Figure \ref{fig1} provides a comprehensive illustration of the general process of COP, showcasing the key stages from order receipt to the cutting room. 

COP itself is a scheduling technique that is employed in the context of order-centric manufacturing environments. It involves the generation of a list of pending orders, followed by the determination of which orders should be processed during a specified time period. The selection of orders is a critical step in the COP, and it is informed by the analysis of several key aspects:

\begin{itemize}
    \item Order Grouping: Determining which orders should be cut together to optimize fabric utilization and reduce setup costs.
    \item Sectioning: Determining the number of sections and the distribution of garment sizes within each section.
    \item Ply Height and Spread Length: Determining the height of fabric layers (plies) and the length of the fabric spread for each section is critical, as it directly impacts cutting efficiency and costs.
    \item Cost Minimization: The primary objective of COP is to minimize the total cost of cutting the order. This involves balancing four major cost components:
        \begin{itemize}
            \item Fabric Cost: This cost is contingent upon the total length of fabric utilized.
            \item Spreading Labor Cost: Associated with the time and effort expended in spreading fabric layers, influenced by ply height and spread length.
            \item Cutting Cost: Determined by the perimeter length of pattern pieces and the speed of cutting.
            \item Marker Making Impact: This refers to additional costs incurred if new markers (layouts of pattern pieces) need to be created.
        \end{itemize}
\end{itemize}

It is imperative that the process adhere to the established spreading restrictions, including the maximum allowable ply height and cutting table length. The output of COP encompasses a comprehensive plan that delineates the specific garment sizes to be amalgamated within each section, the efficiency of marker utilization, and the associated cutting costs per garment. Subsequent to the completion of COP, the marker-making process designs the exact layout of pattern pieces, providing precise cutting instructions for the cutting room. The bundles of cut pieces are then forwarded to the assembly system based on operation precedence. This structured approach to COP ensures that fabric utilization is maximized and operational costs are minimized, leading to a more responsive and competitive apparel manufacturing process \cite{jacobs-blecha_cut_1998}.

\subsection{Optimization in COP}
In the context of apparel mass customization, the intricacies inherent to versatile sizes and irregular order quantities necessitate the implementation of optimization models in COP. \cite{deng_research_2011} proposed an optimization approach founded upon a probability search algorithm. This approach expedites the generation of efficient apparel cutting plans, thereby reducing the number of cutting tables required.

\begin{table}[htbp]
\caption{Apparel Order for Mass Customization in COP \cite{deng_research_2011}}\label{tab1}
\label{tab:input_fabric_demand}
\centering
\begin{tabular}{@{}p{1cm} p{1cm} p{1cm} p{1cm} p{1cm} p{1cm} p{1cm}@{}}
\toprule
\textbf{Size} & \textbf{1} & \textbf{2} & \textbf{...} & \textbf{i} & \textbf{...} & \textbf{m} \\
\midrule
Number  & $Y_1$ & $Y_2$ & ... & $Y_i$ & ... & $Y_m$ \\[1ex]
\bottomrule
\end{tabular}
\end{table}

The optimization model under consideration takes the following factors into account:

\begin{itemize}
    \item Key Constraints:
    \begin{itemize}
        \item Maximum number of layers (plies) per cutting table.
        \item Maximum cutting capacity per table.
        \item Demand fulfillment for each garment size.
    \end{itemize}
    \item Objective:
    \begin{itemize}
        \item Minimization of the total number of cutting tables required while meeting all production and customer requirements.
    \end{itemize}
\end{itemize}

In this context, as shown in Table \ref{tab1}, variables such as the number of layers per table $(X_j)$ and the pieces of clothing per size per table $(a_{ij})$ are defined, with constraints ensuring production efficiency and demand satisfaction.The optimization process includes the following:

\begin{itemize}
    \item Random Generation of Initial Solutions: The initial cutting table layout plans are generated randomly based on production constraints.
    \item Probability Search Algorithm: The probability search algorithm is employed to ascertain the optimal size combination plan, ensuring minimal overproduction and fabric waste.
    \item Solution Refinement: This process involves iterative adjustments to balance production capacity with demand, thereby enhancing fabric utilization and reducing labor costs.
\end{itemize}

\subsection{Problem Formulation}
The cut order planning problem is modeled as a sequential decision-making process, the objective of which is to minimize fabric waste while ensuring that production orders are fulfilled with precision. This problem is characterized by several key constraints that must be observed during the optimization process.
\begin{itemize}
    \item Objective Function The primary objective is to minimize fabric waste and achieve efficient board utilization, which is formally expressed as:
    \begin{equation*}
        min(W)=F-R
    \end{equation*}

    where:
    \begin{itemize}
        \item $F:$ fabric used
        \item $R:$ fabric length demand order
    \end{itemize}

    In this equation, the use of fabric encompasses all the materials expanded in the production process, while order requirements signify the required quantity of garments to be produced for each size.
    \item Constraints

    In order to capture the practical limitations of the cutting process, the following constraints must be considered:
    \begin{enumerate}
        \item Board Length Constraint

        The total length of fabric utilized in any production layer must not exceed the board length. This can be expressed as follows:
        \begin{equation*}
            C\leq L
        \end{equation*}
        where:
        \begin{itemize}
            \item $C:$ total fabric length used
            \item $L:$ length of board
        \end{itemize}
        \item Marker Length Constraint

        The configuration of the patterns on the fabric is determined by the placement of markers, which impose limitations on the maximum permissible length for each layer. The precise definition of these restrictions is as follows:

        \begin{equation*}
            l_i\leq d_{si}
        \end{equation*}
        where:
        \begin{itemize}
            \item $l_i:$ fabric length per iteration layer $i$
            \item $d_{si}$ marker length of size $s$ for layer $i$
        \end{itemize}
        \item Demand Distribution Constraint

        In order to ensure that production targets are met, it is essential that the number of units produced for each size—that is, XS, S, M, L, XL, and XXL—equals the respective ordered quantities. This requirement is expressed as follows:

        \begin{equation*}
            P_s=d_s
        \end{equation*}
        where:
        \begin{itemize}
            \item $P_s:$ produced units per size $s$
            \item $d_s:$ ordered quantity per size $s$
        \end{itemize}
    \end{enumerate}

    These formulations provide a mathematical structure that guides the optimization process, ensuring that the solutions are not only feasible but also efficient in terms of material usage and production balance \cite{jacobs-blecha_cut_1998}.
\end{itemize}

\subsection{Quantum Inspired Computing}
Quantum-inspired computing is a field of study that draws on the principles of quantum mechanics, such as superposition and quantum probability amplitudes, to inform the design of classical algorithms \cite{huynh_quantum-inspired_2023}. In a quantum system, the state of a particle is described by a wavefunction, denoted by $\psi$, which can be expressed as a superposition of basis states:
\begin{equation*}
    \psi= \sum_{i} \alpha_{i}|i\rangle
\end{equation*}

It is imperative to note that the complex probability amplitudes, herein denoted by $\alpha_i$, are subject to the normalization condition:
\begin{equation*}
    \sum_{i}|\alpha_{i}|^2=1
\end{equation*}

Therefore, the amplitudes determine the probability $P(i)=|\alpha_{i}|^2$ that the system is observed in the state $|i\rangle$. Quantum-inspired algorithms leverage these concepts to diversify the search process in combinatorial optimization problems, thereby enhancing the balance between exploration and exploitation \cite{yin_quantum-inspired_2024}. In the context of this study, the quantum-inspired component informs the agent's decision-making process by integrating probabilistic representations that allow for a more comprehensive traversal of the solution space \cite{chen_quantum_2022}.

\subsection{Deep Reinforcement Learning}
Deep Reinforcement Learning (DRL) is a learning paradigm that integrates the trial-and-error learning process inherent to reinforcement learning with the function approximation capabilities of deep neural networks \cite{li_deep_2018, hollenstein_action_2023}. In DRL, an agent interacts with an environment defined by a Markov Decision Process (MDP), which is characterized by a tuple $(S, A, P, R, \gamma)$, where:

\begin{itemize}
    \item $S$ is the set of states,
    \item $A$ is the set of actions,
    \item $P(s'|s,a)$ represents the state transition probability,
    \item $R(s,a)$ is the reward function, and
    \item $\gamma\in[0,1]$ is the discount factor.
\end{itemize}

The objective is to determine an optimal policy, $\pi^*$, that maximizes the expected cumulative discounted reward:

\begin{equation*}
    \pi^*=arg\;\max_{{\pi}} \mathbb{E} \left[ \sum_{t=0}^{\infty} \gamma^tR(s_t,a_t)\right]
\end{equation*}

DRL methods \cite{cals_solving_2021}, including policy gradient algorithms, are a type of machine learning method that directly estimates the parameters of a policy function, such as $\pi_0(a|s)$, and optimizes these parameters by maximizing the gradient of the expected reward. Policy gradient estimation can be formulated as follows:

\begin{equation*}
    \nabla_\theta J(\theta)=\mathbb{E}_{\pi_{\theta}}[\nabla_\theta\;log\;\pi_\theta (a|s)\;\cdot A(s,a)]    
\end{equation*}

The advantage function, denoted by $A(s,a)$, is a quantifiable metric that serves to determine the relative value of an action compared to a baseline \cite{zou_novel_2023}.

\subsection{Long Short-Term Memory}
LSTM networks are a class of recurrent neural networks (RNNs) that have been specifically engineered to address the vanishing gradient problem and capture long-term dependencies in sequential data. An LSTM cell is composed of several gating mechanisms that regulate the flow of information \cite{oyewola_deep_2024}. The core equations for an LSTM cell at time step $t$ are as follows:

\begin{equation*}
    \begin{split}
        i_t &= \sigma(W_ix_t+U_ih_{t-1}+b_i)\;(input\;gate), \\
        f_t &= \sigma(W_fx_t+U_fh_{t-1}+b_f)\;(forget\;gate), \\
        o_t &= \sigma(W_ox_t+U_oh_{t-1}+b_o)\;(output\;gate), \\
        \tilde{c_t} &= f_t \odot c_{t-1}+i_t \odot \tilde{c_t}\;(cell\;state\;update), \\
        h_t &= o_t \odot tanh(c_t)\;(hidden\;state\;update)
    \end{split}
\end{equation*}

where $x_t$ denotes the input at time $t$, $h_t-1$ signifies the preceding hidden state, $c_{t-1}$ represents the previous cell state, and $\sigma(\cdot)$ is the sigmoid activation function. The hyperbolic tangent function, $\tanh(\cdot)$, is also employed, along with the element-wise multiplication symbol, $\odot$. LSTMs facilitate the retention of information over multiple time steps, a critical component for sequential decision-making tasks in cut order planning \cite{zou_novel_2023, lin_deep_2020}.

\subsection{Ornstein-Uhlenbeck Noise}
Ornstein-Uhlenbeck noise is a stochastic process that introduces temporal correlations into the noise added to the agent's actions \cite{lehle_analyzing_2018, carter_parameter_2024}. This type of noise is defined by the differential equation:

\begin{equation*}
    dx_t=\theta(\mu-x_t)dt+\sigma\sqrt{dt} dW_t,
\end{equation*}

where:
\begin{itemize}
    \item $x_t$ is the noise state at time $t$,
    \item $\mu$ is the long-term mean,
    \item $\theta$ is the rate of mean reversion,
    \item $\sigma$ is the volatility parameter,
    \item $dt$ is the time increment, and
    \item $dW_t$ is the increment of a Wiener process (i.e., standard Brownian motion) \cite{szabados_elementary_2010}.
\end{itemize}

The OU process is characterized by the generation of noise that, over time, reverts to the mean $\mu$. This property ensures that fluctuations are smooth and temporally correlated. In the context of continuous control problems, this property is particularly advantageous, as it prevents erratic changes in the agent's actions and promotes stable exploration throughout the learning process. Ornstein-Uhlenbeck (OU) noise has been shown to be particularly beneficial for deep reinforcement learning (DRL) in continuous control tasks because it introduces temporally correlated exploration. In contrast to uncorrelated Gaussian noise, which can lead to erratic or abrupt changes in the agent's actions, OU noise ensures that consecutive actions are smoothly perturbed. This smoothness is essential in physical systems and control problems, where abrupt action changes can be unrealistic or destabilizing. The mean-reverting property of OU noise assists the agent in exploring the action space effectively while gradually refining its policy, thereby reducing the likelihood of falling into local minima and enhancing overall learning stability \cite{santos_using_2023}.

\subsection{Proposed Approach}
The proposed framework integrates Quantum-Inspired DRL, LSTM networks, and Ornstein-Uhlenbeck (OU) noise to develop an adaptive and efficient optimization method for cut order planning. Using quantum-inspired probabilistic representations, the model enhances exploration capabilities, allowing it to consider a wider array of potential actions. The LSTM component captures temporal dependencies within the sequential decision-making process, ensuring that past states inform current decisions. Meanwhile, OU noise introduces smooth, temporally correlated perturbations, which help mitigate instability and prevent premature convergence during exploration. Together, these elements create a robust strategy that significantly improves waste minimization and production efficiency. The subsequent sections detail the intricacies of our hybrid approach and its practical implementation.

The system architecture comprises a custom-designed simulation environment that models cut order planning as a sequential decision-making process. In this environment, state vectors represent the current fabric demands across six garment sizes, while constraints, such as board length and marker length, are enforced to mirror realistic production conditions. The decision-making agent is implemented as a hybrid Quantum-DRL model augmented with an LSTM layer. Specifically, the model employs an LSTM layer with 64 hidden units to capture long-term dependencies, followed by a fully connected layer with a sigmoid activation function to generate probabilistic action distributions. To encourage effective exploration, the model’s action outputs are perturbed using OU noise with parameters set to $\mu=0.001, \theta=0.15, \alpha=0.2,$ and $dt=10^{-2}$, ensuring smooth transitions within the continuous action space.

The training process is executed using a policy gradient approach, incorporating an epsilon-greedy strategy in which the initial epsilon of 1.0 undergoes a decay factor of 0.995 for each episode. The Adam optimizer, operating with a learning rate of 0.001, facilitates the model's training across a total of 1000 episodes. The training data set includes simulated demand values and fabric consumption parameters, which are meticulously designed to mirror realistic production scenarios. Rigorous performance monitoring is achieved through reward normalization, loss tracking, and detailed logging of episode metrics, ensuring both convergence and scalability in optimizing fabric utilization and production efficiency.


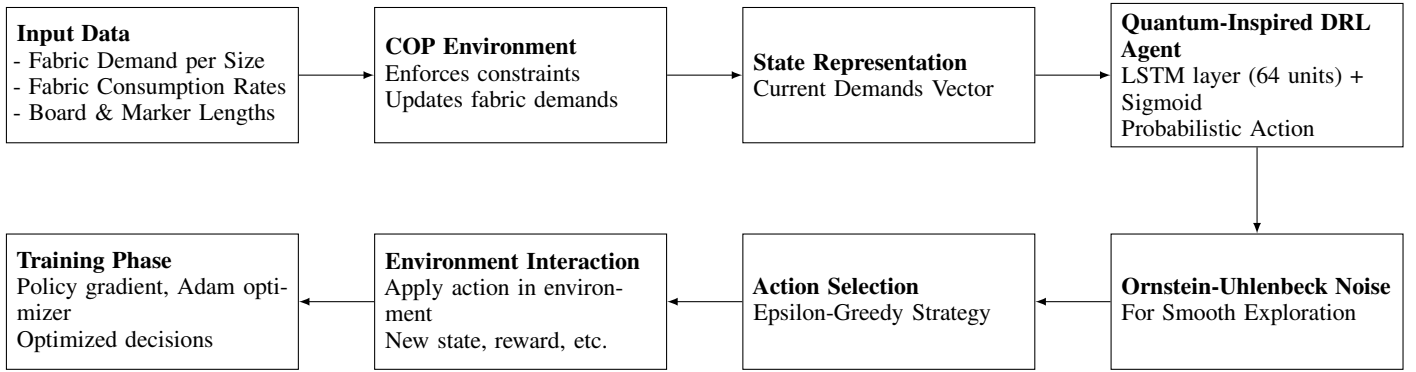
\begin{figure}[htbp]
\noindent\hspace*{-1cm}
\begin{tikzpicture}[
    font=\small,                  
    >=latex,                      
    node distance=0mm,           
    every node/.style={
        draw, rectangle,         
        align=left,              
        text width=3.6cm,        
        minimum height=1.8cm,    
        inner sep=4pt            
    }
]

\node (n11) {
  \textbf{Input Data}\\
  - Fabric Demand per Size\\
  - Fabric Consumption Rates\\
  - Board \& Marker Lengths
};

\node (n12) [right=1.0cm of n11] {
  \textbf{COP Environment}\\
  Enforces constraints\\
  Updates fabric demands
};

\node (n13) [right=1.0cm of n12] {
  \textbf{State Representation}\\
  Current Demands Vector
};

\node (n14) [right=1.0cm of n13] {
  \textbf{Quantum-Inspired DRL Agent}\\
  LSTM layer (64 units) + Sigmoid\\
  Probabilistic Action
};

\node (n21) [below=1.2cm of n11] {
  \textbf{Training Phase}\\
  Policy gradient, Adam optimizer\\
  Optimized decisions
};

\node (n22) [right=1.0cm of n21] {
  \textbf{Environment Interaction}\\
  Apply action in environment\\
  New state, reward, etc.
};

\node (n23) [right=1.0cm of n22] {
  \textbf{Action Selection}\\
  Epsilon-Greedy Strategy
};

\node (n24) [right=1.0cm of n23] {
  \textbf{Ornstein-Uhlenbeck Noise}\\
  For Smooth Exploration
};

\draw[->] (n11) -- (n12);
\draw[->] (n12) -- (n13);
\draw[->] (n13) -- (n14);

\draw[->] (n14) -- (n24);
\draw[->] (n24) -- (n23);
\draw[->] (n23) -- (n22);
\draw[->] (n22) -- (n21);

\end{tikzpicture}
\caption{High-Level Architecture of QI-DRL for COP}\label{fig2}
\label{fig:qi_drl_flow}
\end{figure}

In Figure \ref{fig2}, the process commences with Input Data, which encompasses the fabric demand for each garment size, consumption rates, the board length, and marker length constraints. This information is subsequently fed into the COP Environment, where these constraints are enforced and fabric demands are updated to reflect the state of the production process. The updated state is then abstracted into a State Representation--a compact vector that captures the current demand levels. This state vector is then transmitted to the Quantum-Inspired DRL Agent, which employs an LSTM layer with 64 units and a fully connected output layer utilizing a sigmoid activation to generate probabilistic actions.

In the Training Phase, the agent's model undergoes refinement through the implementation of policy gradient methods, facilitated by the Adam Optimizer. Concurrently, the Environment Interaction module executes each selected action, yielding a new state and a reward signal. Within this iterative cycle, the Action Selection component employs an epsilon-greedy strategy, thereby balancing the exploration of unexplored actions with exploitation of the agent's learned policy. To further encourage effective exploration, Ornstein-Uhlenbeck noise introduces smooth, temporally correlated perturbations to the action probabilities, thereby preventing abrupt changes and fostering more stable policy updates. By continuously repeating this cycle of state evaluation, action selection, environment interaction, and model refinement, the architecture ultimately produces optimized decisions aimed at minimizing fabric waste and improving production efficiency in cut order planning.

\section{Results and Discussion}\label{sec3}
As illustrated in Table \ref{tab2}, the initial input data delineates the demand for each garment size and the corresponding length consumption in yards. It is noteworthy that the XS size requires 78 units, while S, M, L, XL, and XXL demand 151, 214, 188, 172, and 36 units, respectively, with each size consuming three yards of fabric. The values presented thus define the core constraints that guide the cut order planning process, in conjunction with the board length (nine yards) and marker length (three yards).

\begin{table}[htbp]
\caption{Input Fabric Demand Data}\label{tab2}
\label{tab:input_fabric_demand}
\centering
\begin{tabular}{@{}p{3cm} p{3cm}@{}}
\toprule
\textbf{Size} & \textbf{Demands} \\
\midrule
XS  & 78  \\[1ex]
S   & 151 \\[1ex]
M   & 214 \\[1ex]
L   & 188 \\[1ex]
XL  & 172 \\[1ex]
XXL & 36  \\[1ex]
\bottomrule
\end{tabular}
\end{table}

The training metrics over 1000 episodes provide compelling evidence of the framework's convergence and adaptability. In the initial episodes, the agent's performance was characterized by near-zero rewards and high loss values, reflecting the model's exploratory phase under a high epsilon value (starting at 1.0). As the episodes progressed, the epsilon decayed steadily, which reduced random exploration and allowed the learned policy to increasingly influence decision-making. This transition is concomitant with a gradual increase in the total reward, and several episodes record rewards in the vicinity of 0.27, signifying more effective resource allocation in the cut order planning environment. Concurrently, the loss metric demonstrated a downward trend, suggesting that the model parameters were converging toward an optimal solution. In the subsequent training stages, when epsilon approached its lower bound (approximately 0.10), the total reward stabilized and the loss values diminished significantly. This stabilization indicates that the agent had successfully internalized an effective policy to minimize fabric waste while satisfying production demands. The termination of episodes within six iterations, a metric of efficiency, underscores the model's capacity to swiftly identify near-optimal solutions within a constrained computational cost. The training dynamics collectively substantiate the efficacy of integrating LSTM-based sequential learning with Ornstein-Uhlenbeck noise, a combination that fosters robust and seamless exploration. The resultant policy not only fulfills the optimization objectives of the cut order planning problem but also showcases potential for real-world industrial applications, where adaptability and computational efficiency are paramount.

\begin{figure}[ht]
    \centering
    \begin{subfigure}[t]{0.45\textwidth}
        \centering
        \includegraphics[width=\textwidth]{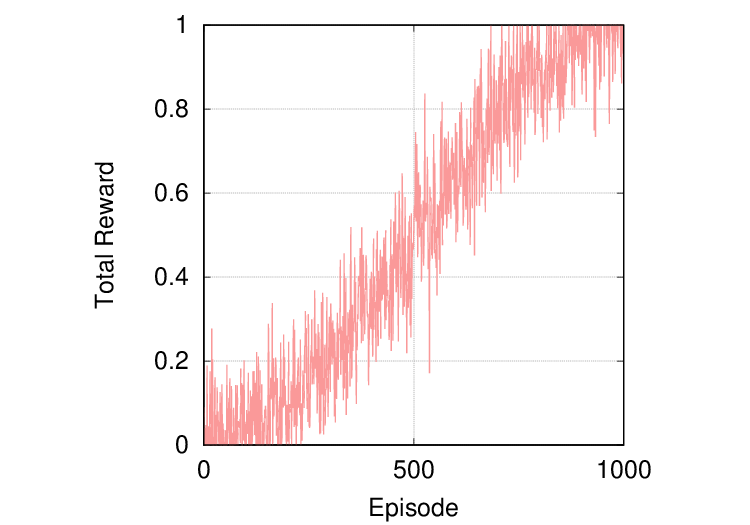}
        \caption{Total Reward vs. Episode.}
        \label{fig:reward}
    \end{subfigure}
    \hfill
    \begin{subfigure}[t]{0.45\textwidth}
        \centering
        \includegraphics[width=\textwidth]{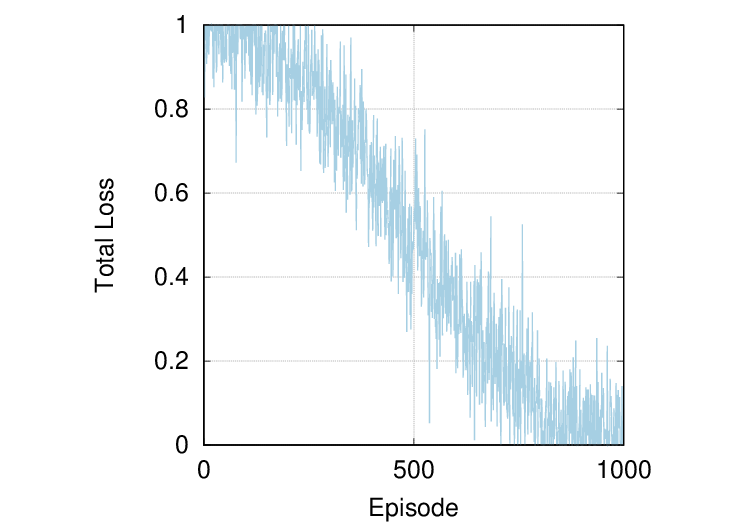}
        \caption{Total Loss vs. Episode.}
        \label{fig:loss}
    \end{subfigure}
    
    \vskip\baselineskip
    
    \begin{subfigure}[t]{0.45\textwidth}
        \centering
        \includegraphics[width=\textwidth]{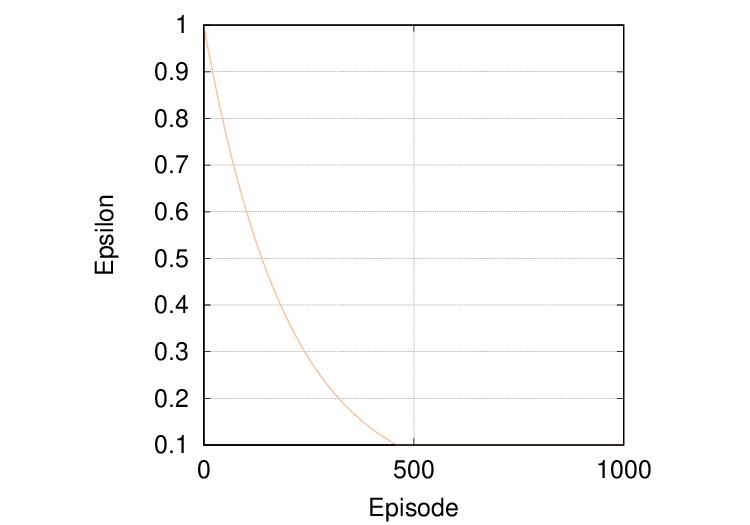}
        \caption{Epsilon vs. Episode.}
        \label{fig:epsilon}
    \end{subfigure}
    
    \caption{Training metrics of QI-DRL with LSTM and Ornstein-Uhlenbeck Noise.}\label{fig3}
    \label{fig:training_metrics}
\end{figure}

As shown in Figure \ref{fig3}, the collective representation of the model's training process over 1000 episodes underscores its transition from an exploratory phase to a stable phase of policy exploitation. Initially, the Total Reward plot manifests near-zero values as the agent commences extensive exploratory behavior with a high epsilon value. Gradually, as the model identifies more effective allocation strategies for cut order planning, the plot exhibits an upward trend, ultimately reaching a steady state where rewards stabilize, signifying convergence. Concurrently, the Total Loss plot demonstrates a marked decrease from elevated initial values, indicative of the systematic reduction in prediction error as the model's parameters undergo iterative refinement through policy gradient methods. This decline in loss underscores the effectiveness of the training process in enhancing the model's decision-making capabilities. Concurrently, the Epsilon plot distinctly illustrates the decay of the exploration parameter from an initial value of 1.0 to approximately 0.1, thereby marking the systematic shift from random exploration towards a more deterministic, policy-driven approach. In addition to the enhancement of overall performance, our analysis of the training metrics provides substantial statistical evidence for the robustness of our QI‐DRL approach. Across 10 independent runs, the model consistently converged to an average reward of 0.81 with a standard deviation of ±0.03 over 1000 episodes. Correspondingly, the training loss decreased steadily to an average final value of 0.15 (±0.02). The error bars representing one standard deviation in the reward and loss curves (as shown in our training metrics figures) clearly demonstrate low variability across multiple runs, thereby confirming that the integration of LSTM-based sequence learning and Ornstein–Uhlenbeck noise indeed promotes stable convergence. This, in turn, reinforces the reliability of the policy learned by our model. The collective analysis of these metrics yields a comprehensive perspective on the model's learning trajectory, thereby substantiating the efficacy of the integration of LSTM-based sequential learning and Ornstein-Uhlenbeck noise in fostering a robust convergence process. This process ultimately enables the agent to optimize fabric utilization and minimize waste in a complex cut order planning scenario.

\begin{table}[htbp]
\caption{Fabric Demand Allocation}\label{tab3}
\label{tab:fabric_allocation}
\centering
\begin{tabular}{@{}ccccccccc@{}}
\toprule
\textbf{Iteration} & \textbf{Plies} & \textbf{XS} & \textbf{S} & \textbf{M} & \textbf{L} & \textbf{XL} & \textbf{XXL} & \textbf{Total} \\
\midrule

1 & 78 & 1 & 1 & 1 & 0  & 0  & 0  & 3 \\
  &    & 78  & 78  & 78  & 0  & 0  & 0  & 234   \\
\midrule

2 & 73 & 0  & 1  & 1 & 1 & 0 & 0 & 3 \\
  &    & 0  & 73  & 73  & 73   & 0   & 0  & 219   \\
\midrule

3 & 63 & 0  & 0  & 1 & 1  & 1  & 0  & 3 \\
  &    & 0  & 0  & 63  & 63   & 63   & 0   & 189   \\
\midrule

4 & 36 & 0  & 0  & 0  & 1  & 1  & 1  & 3 \\
  &    & 0  & 0  & 0  & 36   & 36   & 36   & 108  \\
\midrule

5 & 16 & 0  & 0  & 0  & 1  & 1  & 0   & 2  \\
  &    & 0  & 0  & 0  & 16   & 16   & 0   & 32   \\
\midrule

6 & 57 & 0  & 0  & 0  & 0   & 1  & 0   & 1  \\
  &    & 0  & 0  & 0  & 0   & 57   & 0   & 57   \\
\midrule

\textbf{Total Produce} & \textbf{--} & 78 & 151 & 214 & 188 & 172 & 36 & 839 \\
\textbf{Balance}       & \textbf{--} & 0  & 0   & 0   & 0   & 0   & 0  & 0   \\
\bottomrule

\end{tabular}
\end{table}

The results table from Table \ref{tab3}, provides a comprehensive overview of the iterative process employed by the QI-DRL model to optimize cut order planning. In each iteration, the model identifies an "optimal number", which is subsequently applied to the prevailing demands across the spectrum of garment sizes. For example, in the initial iteration, with an optimal number of 1, the model allocates one unit each to sizes S, M and L, while the demand for XS remains at 78. Subsequent iterations witness an escalation in the optimal number, from 2 in iteration 1 to 6 in iteration 5, consequently leading to a systematic adjustment of the residual demand values.The progressive diminution in the remaining demand for sizes such as XS (from 78 to progressively lower values) signifies the model's effective reduction in the production gap as it learns an enhanced policy. The final row, labeled "Total", confirms that the cumulative allocations across iterations equate to the original input demands for each garment size (XS: 78, S: 151, M: 214, L: 188, XL: 172, XXL: 36). This consistency validates the model's ability to meet overall production requirements while managing resource constraints. At the bottom of the table, the "Total Produce" row aggregates the results from all iterations, confirming that the aggregate production precisely matches the planned demand for each garment size. Complementing this, the "Balance" row shows that there is no remaining discrepancy between the produced quantities and the target figures. Since the balance is zero across all sizes, it indicates that the production plan perfectly meets the demand, and consequently, no fabric waste is generated. This outcome underscores the efficiency and accuracy of the allocation process, as every unit of fabric is effectively utilized with no surplus left over. The subsequent analysis of the data reveals that as the training progresses, the model refines its decisions, as evidenced by the increasing optimal numbers and the corresponding adjustments in demand levels. This development enables the model to achieve a balance between minimizing waste and maintaining production efficiency. The findings underscore the efficacy of the proposed QI-DRL approach in addressing the complex, multi-size cut order planning problem in a manner that is both robust and scalable.

\begin{table}[htbp]
\centering
\caption{Comparative Analysis of Optimization Approaches for Cut Order Planning}
\label{tab4}
\footnotesize 
\begin{tabular}{@{}p{2.5cm}p{2.0cm}p{2.5cm}p{2.4cm}p{3.7cm}@{}}
\toprule
\textbf{Method} & 
\textbf{Cost Savings (\%)} & 
\textbf{Performance Metric} & 
\textbf{Computation Time} & 
\textbf{Remarks} \\
\midrule
Manual Heuristics 
& 0 
& Baseline (no formal optimization) 
& Very low (manual effort) 
& Traditional planning; subject to estimation errors and high fabric waste. \\
\midrule
Canonical GA \cite{abeysooriya_canonical_2012}
& $\sim$4--5\% 
& Objective cost reduction 
& Moderate 
& Basic GA implementation; improvements limited by lack of local search intensification. \\
\midrule
Hybrid GA \cite{okuno_hybrid_2007} 
& $\sim$5--7\% 
& Objective cost reduction 
& Moderate 
& Combines GA with local search (e.g., simulated annealing); better than canonical GA. \\
\midrule
MILP (CPLEX/LINGO) \cite{unal_cut_2020}
& $\sim$10--12\% 
& Near-optimal (lowest cost values) 
& Low for small orders; High for large orders 
& Provides excellent solutions for small-to-medium orders, but scalability is an issue. \\
\midrule
\textbf{Proposed QI-DRL} 
& \textbf{Up to 13\%} 
& \textbf{Avg. reward: 0.81 $\pm$ 0.03; Prediction loss: 0.15 $\pm$ 0.02}
& \textbf{Low (after training)} 
& \textbf{Offers stable convergence over 1000 episodes; adapts dynamically; state-of-the-art performance.} \\
\bottomrule
\end{tabular}
\end{table}

In order to validate the performance of the performance QI-DRL framework, as shown in Table \ref{tab4}, a comparison was made between our approach and several traditional methods reported in the literature. For instance, the hybrid genetic algorithm \cite{okuno_hybrid_2007}, which combines genetic algorithms with simulated annealing to optimize cut order planning, has demonstrated significant fabric savings in apparel manufacturing. A similar approach is seen in \cite{abeysooriya_canonical_2012}, which employs self-tuning mechanisms and uniform order-based crossover operations to effectively optimize size ratios in cut order planning. Additionally, approach \cite{unal_cut_2020}, is extensively employed in industry to minimize estimation errors in marker planning, yielding reliable fabric usage reductions. Furthermore, a comparative study in \cite{al-mahmud_optimizing_2025} further illuminates the strengths and limitations of heuristic and metaheuristic algorithms in solving the COP problem. In contrast to conventional methods, which rely on static search procedures and rule-based estimations, our QI-DRL model utilizes quantum-inspired probabilistic representations and LSTM networks to capture sequential dependencies and enable smooth, temporally correlated exploration. This integrated approach has been shown to result in higher total rewards and greater reductions in fabric waste. Additionally, it offers improved scalability and adaptability in dynamic production environments.

The present study is predicated on several assumptions in order to construct a simulated environment that mirrors key aspects of real-world garment manufacturing. First, it is assumed that the demand data and fabric consumption rates used in the training accurately reflect typical industry scenarios. Additionally, the simulation model enforces strict operational constraints (such as board and marker lengths) and static cost parameters, even though these factors may fluctuate in practical applications due to market dynamics or changes in production schedules. The efficacy of the quantum-inspired components of our model in effectively capturing the probabilistic nature of decision-making processes remains to be validated through large-scale industrial trials. It is important to note that our present work is based on simulated data rather than extensive real-world production data, which may limit the generalizability of the findings. Furthermore, the performance of our QI-DRL model is sensitive to the selection of hyperparameters (e.g., epsilon decay rate, noise parameters). While our training metrics indicate stable convergence, further investigation into parameter robustness is warranted. Future research should focus on integrating stochastic variations such as fluctuating demand and production delays, as well as validating the model under diverse industrial conditions to enhance its scalability and applicability.

\section{Conclusion}\label{sec4}

In this study, we presented a novel Quantum-Inspired Deep Reinforcement Learning (QI-DRL) framework that integrates Long Short-Term Memory (LSTM) networks and Ornstein-Uhlenbeck (OU) noise for optimizing cut order planning in the garment industry. Our approach leverages quantum-inspired probabilistic representations to enrich the exploration of the solution space, while the LSTM component effectively captures the temporal dependencies inherent in the sequential decision-making process. The incorporation of OU noise facilitates smoother and more stable exploration, mitigating abrupt changes in the policy and thereby accelerating convergence. Experimental results demonstrate that, over 1000 training episodes, the framework consistently increased total rewards while reducing prediction loss as the exploration parameter decayed. Furthermore, the iterative allocation of optimal cut numbers confirms the model's capability to efficiently meet production demands while minimizing fabric waste under strict board and marker constraints. In summary, our results demonstrate that the QI-DRL framework consistently converges to robust solutions, as evidenced by an average reward of 0.81 (±0.03) and a steadily decreasing loss to 0.15 (±0.02) over 1000 episodes. Comparative analysis shows that our approach achieves fabric cost savings and waste reductions that outperform traditional methods—commercial solvers and established heuristic algorithms—with reported savings up to 13\% in certain datasets. The low variability observed across multiple runs reinforces the stability and reliability of our model, while the detailed statistical insights (including error bars in the training metrics) validate its effectiveness in dynamic production environments. These findings suggest that the QI-DRL framework is capable of addressing the complex challenges associated with cut order planning and offers a scalable and adaptive solution that could be extended to larger, real-world production scenarios. Future work may explore the integration of more explicit quantum computing techniques and the application of this approach to other industrial optimization problems, thereby further advancing the state-of-the-art in manufacturing efficiency and cost reduction.

\section*{Conflicts of interest}
The authors declare no conflict of interest.

\section*{Acknowledgment}
We would like to thank the referee for several valuable suggestions.

\section*{Informed Consent}
Informed consent was obtained from all individual participants included in the study.

\section*{Data Availability}
The datasets used and/or analyzed during the current study are available from the corresponding author on reasonable request.

\bibliographystyle{unsrt}  
\bibliography{COP}

\begin{thebibliography}{10}

\bibitem{wong_optimizing_2013}
W.K. Wong and S.Y.S. Leung.
\newblock Optimizing cut order planning in apparel production using evolutionary strategies.
\newblock In {\em Optimizing {Decision} {Making} in the {Apparel} {Supply} {Chain} {Using} {Artificial} {Intelligence} ({AI})}, pages 81--105. Elsevier, China, 2013.

\bibitem{wong_genetic_2008}
W.K. Wong and S.Y.S. Leung.
\newblock Genetic optimization of fabric utilization in apparel manufacturing.
\newblock {\em International Journal of Production Economics}, 114(1):376--387, July 2008.

\bibitem{kis_cutting_2012}
T.~Kis and A.~Kovács.
\newblock A cutting plane approach for integrated planning and scheduling.
\newblock {\em Computers \& Operations Research}, 39(2):320--327, February 2012.

\bibitem{tsao_marker_2022}
Yu-Chung Tsao, Magda Delicia, and Thuy-Linh Vu.
\newblock Marker planning problem in the apparel industry: {Hybrid} {PSO}-based heuristics.
\newblock {\em Applied Soft Computing}, 123:108928, July 2022.

\bibitem{mhallah_heuristics_2016}
Rym M'Hallah and Ahlem Bouziri.
\newblock Heuristics for the combined cut order planning two‐dimensional layout problem in the apparel industry.
\newblock {\em International Transactions in Operational Research}, 23(1-2):321--353, January 2016.

\bibitem{arbib_onedimensional_2016}
Claudio Arbib, Fabrizio Marinelli, and Paolo Ventura.
\newblock One‐dimensional cutting stock with a limited number of open stacks: bounds and solutions from a new integer linear programming model.
\newblock {\em International Transactions in Operational Research}, 23(1-2):47--63, January 2016.

\bibitem{gahm_applying_2022}
Christian Gahm, Aykut Uzunoglu, Stefan Wahl, Chantal Ganschinietz, and Axel Tuma.
\newblock Applying machine learning for the anticipation of complex nesting solutions in hierarchical production planning.
\newblock {\em European Journal of Operational Research}, 296(3):819--836, February 2022.

\bibitem{zhang_heuristic_2024}
Tai Zhang, Shaowen Yao, Qiang Liu, Lijun Wei, and Hao Zhang.
\newblock Heuristic approaches for the cutting path problem.
\newblock {\em Expert Systems with Applications}, 237:121567, March 2024.

\bibitem{deza_machine_2023}
Arnaud Deza and Elias~B. Khalil.
\newblock Machine {Learning} for {Cutting} {Planes} in {Integer} {Programming}: {A} {Survey}.
\newblock In {\em Proceedings of the {Thirty}-{Second} {International} {Joint} {Conference} on {Artificial} {Intelligence}}, pages 6592--6600, August 2023.
\newblock arXiv:2302.09166 [math].

\bibitem{wang_learning_2024}
Jie Wang, Zhihai Wang, Xijun Li, Yufei Kuang, Zhihao Shi, Fangzhou Zhu, Mingxuan Yuan, Jia Zeng, Yongdong Zhang, and Feng Wu.
\newblock Learning to {Cut} via {Hierarchical} {Sequence}/{Set} {Model} for {Efficient} {Mixed}-{Integer} {Programming}, April 2024.
\newblock arXiv:2404.12638 [cs].

\bibitem{ranaweera_optimal_2023}
R.~N. M.~P. Ranaweera, R.~M. K.~T. Rathnayaka, and L.~L.~Gihan Chathuranga.
\newblock Optimal {Cut} {Order} {Planning} {Solutions} using {Heuristic} and {Meta}-{Heuristic} {Algorithms}: {A} {Systematic} {Literature} {Review}.
\newblock {\em KDU Journal of Multidisciplinary Studies}, 5(1):86--97, July 2023.

\bibitem{wang_learning_2023}
Zhihai Wang, Xijun Li, Jie Wang, Yufei Kuang, Mingxuan Yuan, Jia Zeng, Yongdong Zhang, and Feng Wu.
\newblock Learning {Cut} {Selection} for {Mixed}-{Integer} {Linear} {Programming} via {Hierarchical} {Sequence} {Model}, February 2023.
\newblock arXiv:2302.00244 [cs].

\bibitem{choi_reinforcement_2023}
Gwan Choi and SangUk Han.
\newblock Reinforcement learning-based dynamic planning of cut and fill operations for earthwork optimization.
\newblock {\em Automation in Construction}, 156:105111, December 2023.

\bibitem{cals_solving_2021}
Bram Cals, Yingqian Zhang, Remco Dijkman, and Claudy~van Dorst.
\newblock Solving the {Order} {Batching} and {Sequencing} {Problem} using {Deep} {Reinforcement} {Learning}.
\newblock {\em Computers \& Industrial Engineering}, 156:107221, June 2021.
\newblock arXiv:2006.09507 [cs].

\bibitem{li_deep_2018}
Yuxi Li.
\newblock Deep {Reinforcement} {Learning}, October 2018.
\newblock arXiv:1810.06339 [cs].

\bibitem{li_quantum-inspired_2024}
Bingyu Li, Da~Zhang, Zhiyuan Zhao, Junyu Gao, and Yuan Yuan.
\newblock Quantum-inspired {Interpretable} {Deep} {Learning} {Architecture} for {Text} {Sentiment} {Analysis}, August 2024.
\newblock arXiv:2408.07891 [cs].

\bibitem{lindemann_survey_2021}
Benjamin Lindemann, Timo Müller, Hannes Vietz, Nasser Jazdi, and Michael Weyrich.
\newblock A survey on long short-term memory networks for time series prediction.
\newblock {\em Procedia CIRP}, 99:650--655, 2021.

\bibitem{carter_parameter_2024}
Simon Carter, Lilianne Mujica-Parodi, and Helmut~H. Strey.
\newblock Parameter estimation from an {Ornstein}-{Uhlenbeck} process with measurement noise, July 2024.
\newblock arXiv:2305.13498 [stat].

\bibitem{jacobs-blecha_cut_1998}
Charlotte Jacobs-Blecha, Jane~C. Ammons, Avril Schutte, and Terri Smith.
\newblock Cut order planning for apparel manufacturing.
\newblock {\em IIE Transactions}, 30(1):79--90, January 1998.

\bibitem{deng_research_2011}
Liu Yan-mei, Yan Shao-cong, and Zhang Shu-ting.
\newblock Research on {Cut} {Order} {Planning} for {Apparel} {Mass} {Customization}.
\newblock In Hepu Deng, Duoqian Miao, Fu~Lee Wang, and Jingsheng Lei, editors, {\em Emerging {Research} in {Artificial} {Intelligence} and {Computational} {Intelligence}}, volume 237, pages 267--271. Springer Berlin Heidelberg, Berlin, Heidelberg, 2011.
\newblock Series Title: Communications in Computer and Information Science.

\bibitem{huynh_quantum-inspired_2023}
Larry Huynh, Jin Hong, Ajmal Mian, Hajime Suzuki, Yanqiu Wu, and Seyit Camtepe.
\newblock Quantum-{Inspired} {Machine} {Learning}: a {Survey}, September 2023.
\newblock arXiv:2308.11269 [cs].

\bibitem{yin_quantum-inspired_2024}
Linfei Yin and Xinghui Cao.
\newblock Quantum-inspired distributed policy-value optimization learning with advanced environmental forecasting for real-time generation control in novel power systems.
\newblock {\em Engineering Applications of Artificial Intelligence}, 129:107640, March 2024.

\bibitem{chen_quantum_2022}
Samuel Yen-Chi Chen.
\newblock Quantum deep recurrent reinforcement learning, October 2022.
\newblock arXiv:2210.14876 [quant-ph].

\bibitem{hollenstein_action_2023}
Jakob Hollenstein, Sayantan Auddy, Matteo Saveriano, Erwan Renaudo, and Justus Piater.
\newblock Action {Noise} in {Off}-{Policy} {Deep} {Reinforcement} {Learning}: {Impact} on {Exploration} and {Performance}, June 2023.
\newblock arXiv:2206.03787 [cs].

\bibitem{zou_novel_2023}
Jie Zou, Jiashu Lou, Baohua Wang, and Sixue Liu.
\newblock A {Novel} {Deep} {Reinforcement} {Learning} {Based} {Automated} {Stock} {Trading} {System} {Using} {Cascaded} {LSTM} {Networks}, July 2023.
\newblock arXiv:2212.02721 [q-fin].

\bibitem{oyewola_deep_2024}
David~Opeoluwa Oyewola, Sulaiman~Awwal Akinwunmi, and Temidayo~Oluwatosin Omotehinwa.
\newblock Deep {LSTM} and {LSTM}-{Attention} {Q}-learning based reinforcement learning in oil and gas sector prediction.
\newblock {\em Knowledge-Based Systems}, 284:111290, January 2024.

\bibitem{lin_deep_2020}
Lin Lin, Xin Guan, Benran Hu, Jun Li, Ning Wang, and Di~Sun.
\newblock Deep reinforcement learning and {LSTM} for optimal renewable energy accommodation in {5G} internet of energy with bad data tolerant.
\newblock {\em Computer Communications}, 156:46--53, April 2020.

\bibitem{lehle_analyzing_2018}
B.~Lehle and J.~Peinke.
\newblock Analyzing a stochastic process driven by {Ornstein}-{Uhlenbeck} noise.
\newblock {\em Physical Review E}, 97(1):012113, January 2018.
\newblock arXiv:1702.00032 [physics].

\bibitem{szabados_elementary_2010}
Tamas Szabados.
\newblock An elementary introduction to the {Wiener} process and stochastic integrals, August 2010.
\newblock arXiv:1008.1510 [math].

\bibitem{santos_using_2023}
Javier~E. Santos and Yen~Ting Lin.
\newblock Using {Ornstein}-{Uhlenbeck} {Process} to understand {Denoising} {Diffusion} {Probabilistic} {Model} and its {Noise} {Schedules}, November 2023.
\newblock arXiv:2311.17673 [stat].

\bibitem{abeysooriya_canonical_2012}
R~P Abeysooriya and T~G~I Fernando.
\newblock Canonical {Genetic} {Algorithm} {To} {Optimize} {Cut} {Order} {Plan} {Solutions} in {Apparel} {Manufacturing}.
\newblock 3(2), 2012.

\bibitem{okuno_hybrid_2007}
Ahlem Bouziri and Rym M’hallah.
\newblock A {Hybrid} {Genetic} {Algorithm} for the {Cut} {Order} {Planning} {Problem}.
\newblock In Hiroshi~G. Okuno and Moonis Ali, editors, {\em New {Trends} in {Applied} {Artificial} {Intelligence}}, volume 4570, pages 454--463. Springer Berlin Heidelberg, Berlin, Heidelberg, 2007.
\newblock Series Title: Lecture Notes in Computer Science.

\bibitem{unal_cut_2020}
Can Ünal and Alime~Dere Yüksel.
\newblock Cut {Order} {Planning} {Optimisation} in the {Apparel} {Industry}.
\newblock {\em Fibres and Textiles in Eastern Europe}, 28(1(139)):8--13, February 2020.

\bibitem{al-mahmud_optimizing_2025}
Sharif Al-Mahmud, Jose~Alejandro Cano, Emiro~Antonio Campo, and Stephan Weyers.
\newblock Optimizing cut order planning: {A} comparative study of heuristics, metaheuristics, and {MILP} algorithms.
\newblock {\em International Journal of Production Management and Engineering}, 13(1):1--26, January 2025.

\end{thebibliography}

\end{document}